\documentclass[10pt,twocolumn,letterpaper]{article}

\usepackage{cvpr}
\usepackage{times}
\usepackage{epsfig}
\usepackage{graphicx}
\usepackage{amsmath}
\usepackage{amssymb}
\usepackage{comment}
\usepackage{color}
\usepackage{threeparttable}

\usepackage[breaklinks=true,bookmarks=false]{hyperref}

\cvprfinalcopy 

\def\cvprPaperID{3667} 

\ifcvprfinal\fi
\begin{document}

\title{Semantically Aligned Bias Reducing Zero Shot Learning}

\author{Akanksha Paul\\
Indian Institute of Technology\\
Ropar\\
{\tt\small akanksha.paul@iitrpr.ac.in}
\and
Naraynan C Krishnan\\
Indian Institute of Technology\\
Ropar\\
{\tt\small ckn@iitrpr.ac.in}
\and
Prateek Munjal\\
Indian Institute of Technology\\
Ropar\\
{\tt\small 2017csm1009@iitrpr.ac.in}
}

\maketitle
\thispagestyle{empty}
\begin{abstract}
   Zero shot learning (ZSL) aims to recognize unseen classes by exploiting semantic relationships between seen and unseen classes. Two major problems faced by ZSL algorithms are the hubness problem and the bias towards the seen classes. Existing ZSL methods focus on only one of these problems in the conventional and generalized ZSL setting. In this work, we propose a novel approach, Semantically Aligned Bias Reducing (SABR) ZSL, which focuses on solving both the problems. It overcomes the hubness problem by learning a latent space that preserves the semantic relationship between the labels while encoding the discriminating information about the classes.  Further, we also propose ways to reduce bias of the seen classes through a simple cross-validation process in the inductive setting and a novel weak transfer constraint in the transductive setting.  Extensive experiments on three benchmark datasets suggest that the proposed model significantly outperforms existing state-of-the-art algorithms by $\sim$1.5-9\% in the conventional ZSL setting and by $\sim$2-14\% in the generalized ZSL for both the inductive and transductive settings.
\end{abstract}

\section{Introduction}
In recent years, deep learning has achieved state-of-the-art performance across a wide range of computer vision tasks such as image classification task \cite{imagenet}. However, these deep learning methods rely on enormous amount of labeled data which is scarce for dynamically emerging objects. Practically, it is unrealistic to annotate everything around us. Thus, making the conventional object classification methods infeasible.  
In this work, we focus on the extreme case when there is no labeled data, \ie, Zero-shot learning (ZSL), where the task is to recognize the unseen class instances by relying on the labeled set of seen classes. ZSL assumes that the semantic label embeddings of both seen and unseen classes are known apriori. ZSL thus learns to identify unseen classes by leveraging the semantic relationship between seen and unseen classes.

On the basis of data available during the training phase, ZSL can be divided into two categories: \textbf{Inductive} and \textbf{Transductive ZSL}. In inductive ZSL \cite{devise,lampert2014attribute,semanticae,sje,embarrassinglyzsl,latem,dem,preservingsemantic,DBLP:FeatureGenNetworks}, we are provided with the labeled seen class instances and the semantic embedding of unseen class labels during training. While in transductive ZSL \cite{gfzsl,dsrl}, in addition to the labeled seen class data and the semantic embedding of all labels, we are also provided with the unlabeled instances of unseen classes data. ZSL can also be categorized into \textbf{Conventional} and \textbf{Generalized ZSL} depending on the data that is presented to the model during the testing phase. In Conventional ZSL,  data emerges only from unseen classes at test time. While Generalized ZSL \cite{chao2016empirical} is a more realistic setting where the data during testing comes from both seen and unseen classes.  

Generally, ZSL approaches project the seen and unseen class data into a latent space that is robust for learning unseen class labels. One approach is to learn a latent space that is aligned towards the semantic label embedding \cite{devise,semanticae,lampert2014attribute,embarrassinglyzsl,cmtzerong,latem}. The input data is transformed into this latent space for learning the classification models over seen and unseen classes. This approach leads to the well known \textbf{hubness problem} \cite{DBLP:hubness_prob,radovanovic2010hubs,dinu2014improving} where the transformed data become hubs for the nearby class embeddings leading to performance deterioration in both conventional and generalized ZSL. To alleviate the hubness problem, the other approaches \cite{dem,textzsl,zerozhang,DBLP:hubness_prob} learn a latent visual space for recognizing the seen class labels by aligning the semantic class embeddings towards this latent space. Irrespective of the latent space for transforming the data, there is an inherent bias in the model towards seen classes, which we refer to as the \textbf{bias problem}. Due to this bias the models generally perform poorly on unseen classes. 

Existing ZSL methods focus on addressing only one of these problems. In this work, we propose a novel method - \textbf{Semantically Aligned Bias Reducing (SABR)} ZSL to alleviate both the hubness and bias problems. We propose two versions of SABR - SABR-I and SABR-T for the inductive and the transductive ZSL settings respectively. Both these versions have a common first step that learns an intermediate representation for the seen and unseen class data. 
This intermediate latent space is learned to preserve both the semantic relationship between class embeddings and discriminating information among classes through a novel loss function. 

After having learned the optimal latent space, both SABR-I and SABR-T learn generative adversarial networks (GAN) to generate the latent space representations. Specifically, SABR-I learns a conditional Wasserstein GAN for generating the latent space representations for the seen classes using only the seen class embeddings. 
As the label embeddings of seen and unseen classes exhibit semantic relationships that are being learned in first step, we utilize the generative network to synthesize unseen class representations for learning a classification model for ZSL and GZSL. Given that we only have labeled data for the seen classes, SABR-I reduces the bias by early stopping the training of conditional WGAN through a simulated ZSL problem induced on the seen class data.

SABR-T goes further to learn a different GAN for generating latent space instances for the unseen classes. This network is learned to minimize the marginal probability difference between the true latent space representations of the unlabeled unseen class instances and the synthetically generated representations. Further, the conditional probability distribution of the latent space representations given the semantic labels are weakly transferred from the conditional WGAN learned by SABR-I for the seen class labels. Specifically, we learn a Wasserstein GAN\cite{DBLP:wgan} for the unseen classes by constraining the amount of transfer from the seen classes as learned by SABR-I. 
Overall, the major contributions of the paper are as follows 
\begin{itemize}
    \item We propose a novel two-step solution for zero shot learning. In the first step, an appropriate latent space is learned by fine tuning a pre-trained model with semantic embedding to reduce the hubness problem. In the second step, generators for synthesizing unseen class representations are learned, whose bias towards the seen class is reduced by using an early stopping criterion in the inductive setting and a weak transfer criteria for the transductive setting.
    \item We introduce a loss function, which ensures that the embedding space is discriminative and semantically aligned with the semantic class embeddings. A novel adversarial generative transfer is proposed that tries to minimize both the conditional and marginal distributions of seen and unseen classes. 
    \item Empirical evaluation across all the zero-shot learning datasets suggests that the proposed approach outperforms the state-of-the-art performance in both conventional and generalized ZSL in both the inductive and transductive settings.
\end{itemize}
\begin{figure*}[t]
\centering
\includegraphics[width=0.8\textwidth]{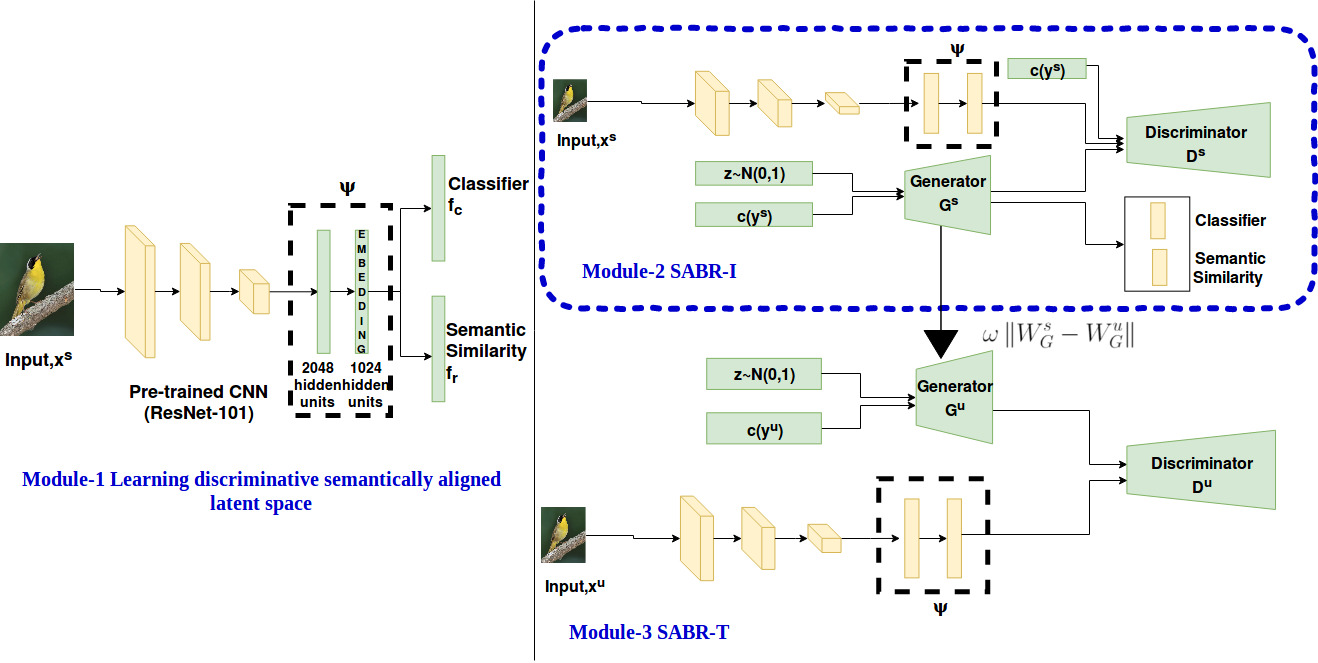}
\caption{[Best viewed in color] An illustration of the proposed Semantically Aligned Bias Reducing (SABR) model.} 
\label{fig:1}
\end{figure*}
\section{Related Work}
Zero-shot learning (ZSL) has been a well studied area in recent years. Early ZSL approaches \cite{devise,semanticae,sje,embarrassinglyzsl,cmtzerong,latem} utilized the semantic label space for projecting the seen and unseen instances. DEVISE \cite{devise}, ALE \cite{ale} and SJE \cite{sje} learned bi-linear compatibility functions to model the relationship between visual and semantic space. ESZSL \cite{embarrassinglyzsl} added a regularizer to the bi-linear compatibility functions that bounded the norm of projected features and semantic attributes. All these methods were constrained by learning linear functions and was overcome by LATEM \cite{latem} and CMT \cite{cmtzerong} which learned non-linear functions.
Zhang \etal \cite{dem} were the first to demonstrate the hubness problem and suggested to use an intermediate visual space for projecting the seen and unseen class instances. 
Zhang \etal \cite{dem} and Ba \etal \cite{textzsl} transform both the semantic and visual space to a joint embedding space in which the visual representations are closer to their respective semantic representations. Annadani \etal \cite{preservingsemantic} focused on utilizing the semantic structure while maintaining separability of classes. Our approach for the inductive ZSL setting (SABR-I) improves over the work of Annadani \etal \cite{preservingsemantic}, as in SABR-I we not only preserve semantic relations in visual space but also reduce the bias of seen classes. 

Among the transductive ZSL approaches, Song \etal \cite{qfzsl} leverage the conditional seen class data with unlabelled unseen class data to learn an unbiased latent embedding space. The bias towards seen classes is reduced by forcing a uniform prior over the output of the classifier for unseen class instances. Our approach differs from QFSL primarily in two ways. Firstly, we reduce the bias in both the inductive and transductive versions of our model. In the transductive version, SABR-T reduces the bias without enforcing the uniform prior on the output as this will reduce the unseen class conditional information in the latent space. Secondly, learning an optimal latent space helps to mitigate the hubness problem.

There also exists work on generative modeling for ZSL \cite{gfzsl,bucher2017generating,DBLP:FeatureGenNetworks}. Verma \etal \cite{gfzsl} model each class-conditional distribution as a Gaussian distribution whose parameters are learned by seen classes. It then predicts the parameters of class-conditional distribution for unseen classes. They further extend this work to incorporate the unseen class data and report results in transductive setting. 
Bucher \etal \cite{bucher2017generating} and Xian \etal \cite{DBLP:FeatureGenNetworks} generate pseudo instances of unseen classes by training a conditional generator for the seen classes. Our proposed approach for the inductive setting SABR-I differs from these approaches, as we learn a discriminative embedding space that preserves semantic relations which alleviates hubness problem. Further, we leverage the unlabeled unseen class data to reduce the bias of seen classes. 
\section{Methodology}
\subsection{Problem Definition}
Let $D_{s}=\{x^s_i, y^s_i, c(y^s_i), i=1,2....N_s\}$ represent the set of seen class data instances, where $x^s_i$ denotes the $i^{th}$ seen class instance with the corresponding class label $y^s_i \in S$ (the set of seen classes). The semantic label embedding for each $y^{s} \in  S$ is denoted by $c(y^{s})$. In the inductive ZSL setting, we are provided only with the set of unseen labels $y^{u} \in U$ and the corresponding semantic label embedding $c(y^{u})$. There is no overlap between the seen and unseen classes i.e., $S \cap U = \emptyset$. In the transductive ZSL setting, we also have unlabeled unseen class data represented by $D_{u}=\{x^{u}_{i}, y^u, c(y^u), i=1,2....N_{u}\}$ where $x^{u}_{i}$ is the $i^{th}$ unseen class instance. As the unseen class dataset is unlabeled, we do not have the labels $y^{u}_{i}$ of $x^{u}_{i}$.  The goal of conventional ZSL is to predict the label for each $x^u_{i} \in U$. In the generalized ZSL setting, the goal is to predict the label of a test sample, where the test sample can belong to either seen or unseen class. 
\subsection{Semantically Aligned Bias Reducing ZSL}
In this section, we present our proposed two-tier model, SABR-I, for inductive ZSL (and a three-tier model, SABR-T for transductive ZSL) as shown in figure \ref{fig:1}. 

\subsubsection{Learning the Optimal Latent Space}
In the first step of SABR-I and SABR-T, we learn a latent space $\Psi$ that preserves the semantic relations between classes while also learning the discriminating information for recognizing the classes. The semantic relations are essential as the learned latent space is later used for generating synthetic instances of unseen classes. The discriminating information is useful for learning the classifier and thus mitigating the hubness effect. We use pre-trained deep network Resnet-101 to extract features from the seen and unseen class images. For simplicity henceforth $x_i^s$ and $x_i^u$ refer to the features extracted from the pre-trained deep embedding models. 
These features are then used to learn a transformation, $\psi()$, that projects the seen and unseen class instances onto the latent space $\Psi$. $\psi()$ is modeled as a two layer fully connected network. The latent representations $\psi(x_i^s)$ are used to simultaneously learn a classifier ($f_c$) (for learning the discriminating information) and a regressor ($f_r$) (for preserving the semantic relationships among the labels). The classifier $f_c$ outputs the one hot encoding of the class label of the instance and thus is trained minimizing the cross entropy loss, 
\begin{equation}\label{eq:1}
\mathcal{L}_{C} = -\frac{1}{N_{s}}\sum_{i=1}^{N_{s}} \mathcal{L}(y^{s}_{i}, f_c(\psi(x^s_i))
\end{equation}
where $\mathcal{L}$ is cross entropy loss between true and predicted labels of seen class instance $x_i^s$.

The semantic relationships between the labels are preserved by ensuring that the output of the regressor $f_r$ on the embedding of a seen instance $\psi(x^s_i)$  is closely related to the corresponding semantic embedding $c(y^s_i)$. We propose to use a similarity based cross-entropy loss, as defined in the equation below, between the predicted label embeddings of the regressor and the true semantic label embedding. 
\begin{equation}\label{eq:2}
\mathcal{L}_{S}=-\sum_{i=1}^{N_{s}} \log\frac{\exp(\langle f_r(\psi(x_i^s)),c(y_i^s)\rangle)}{ \Sigma_{y^s\in S}\exp(\langle f_r(\psi(x_i^s)),c(y_i^s)\rangle)}
\end{equation}
where $\langle f_r(\psi(x_i^s)),c(y_i^s)\rangle$ refers to the similarity between predicted label embedding, $f_r(\psi(x_i^s))$, of each source instance $x^{s}_{i}$ and its true semantic label embedding $c(y^{s}_{i})$. This loss function ensures that the predicted label embeddings for all seen class instances belonging to a specific label form a  cluster around the true semantic label embedding. The similarity could be defined using any measure such as Euclidean distance, cosine similarity or dot product. 

The transformation function $\psi()$, as well as the classifier $f_c$ and the regressor $f_r$ are learned simultaneously by jointly minimizing the loss functions represented in equations \ref{eq:1} and \ref{eq:2} weighted by the factor $\gamma$
\begin{equation}\label{eq:3}
 \mathcal{L}_{FS} = {\text{min}}~\mathcal{L}_{C}+ \gamma*\mathcal{L}_{S}
\end{equation}

Thus at the end of step 1, both the versions of SABR learn the transformation $\psi$ to the latent space that is optimal in the sense that it possesses the discriminative information for classification and encodes the semantic relationship between the labels.

\subsubsection{Bias Reducing Generator Network for SABR-I}
The objective of inductive ZSL is to learn a classifier that can predict the labels of unseen class instances. As we do not have training instances of unseen classes, following the approach of Xian \etal \cite{DBLP:FeatureGenNetworks}, we learn a generator network that can generate synthetic unseen class instances as illustrated in the module 2 of Figure \ref{fig:1}

Given the seen class embeddings $\psi(x_i^s)\in \Psi$, we first learn a conditional generator $G^s: \langle z,c(y^{s}) \rangle \to \Psi$. The generator takes as input a random noise vector $z$ and a semantic label embedding $c(y^{s})$ and outputs an instance $\tilde{x}^s$ in the latent space $\Psi$. As we know the labels associated with each seen class training instance, we train the conditional generator using the Wasserstein adversarial loss defined by
\begin{equation}
\begin{split}
\mathcal{L}^s_G &=\mathop{{}\mathbb{E}}[D^s(\psi(x^{s}),c(y^{s}))]-\mathop{{}\mathbb{E}}[D^s(\tilde{x}^s,c(y^{s}))] \\ &-\lambda\mathop{{}\mathbb{E}}
[(\left\lVert\nabla_{\hat{x}^s}D^s(\hat{x}^s,c(y^s))\right\rVert-1)^2]
\end{split}
\end{equation}
where, $D^s$ is the seen class conditional discriminator whose input is the seen class label embedding ($c(y^s)$) and the latent space instance ($\psi(x^s)$), $\hat{x}^s =\alpha \psi(x^s)+(1-\alpha)\tilde{x}^s$ with $\alpha \sim U(0,1)$ and $\lambda$ is the gradient penalty coefficient. Thus, the objective for discriminator and generator pair is to 

\begin{equation}
    \underset{G^s}{\text{min}}\, \underset{D^s}{\text{max}}\,  \mathcal{L}^s_G
\end{equation}
We further want to encourage the generator to synthesize latent space embeddings of seen classes that are discriminative and encode the semantic similarity between the label embeddings. We achieve this by incorporating the loss functions defined in equation \ref{eq:1} and \ref{eq:2} to the overall optimization objective of the generator. We use the pre-trained classifier $f_c$ and regressor $f_r$ from the previous step while training the generator $G^s$. Thus the overall loss function for the generator-discriminator network can be defined as   
\begin{equation}
    \underset{G^s}{\text{min}}\,  \underset{D^s}{\text{max}}\,  \mathcal{L}^s_G+\beta ( \mathcal{L}_C+ \gamma \mathcal{L}_S)
\end{equation}

This generator is then used to synthesize the latent space representations for the unseen classes. The semantic label embeddings encode relationships between the labels and therefore we expect the generator to synthesize meaningful latent representations of the unseen classes. However, the generator can be overly biased towards the seen classes due to the training set that is presented to it. This bias is mitigated using the principle of early stopping during training of the generator. The number of training epochs required to achieve the best performance is determined through a simple cross-validation set up on the seen classes.

\subsubsection{Bias Reducing Generator Network for SABR-T}
In the transductive setting the training process can benefit from modeling the unlabeled unseen class data. In particular, we model the marginal probability distribution of the unseen class unlabeled data via a GAN. We first obtain the latent representations of unseen class data $x^u$ by transforming them using the function $\psi()$. Now, given the latent space representations of the unseen class instances $\psi(x^u)$, we learn a generator $G^u: \langle z,c(y^{u}) \rangle \to \Psi$ that takes noise z and semantic vector $c(y^{u})$ as the input and outputs a synthetic instance $\tilde{x}^u$ in the latent space. $G^u$ is trained as a conditional generator using the Wasserstein adversarial loss defined as follows 
\begin{equation}
\begin{split}
\mathcal{L}^u_G &=\mathop{{}\mathbb{E}}[D^u(\psi(x^{u}))]-\mathop{{}\mathbb{E}}[D^u(\tilde{x}^u)] \\ &-\lambda\mathop{{}\mathbb{E}}
[(\left\lVert\nabla_{\hat{x}^u}D^u(\hat{x}^u)\right\rVert-1)^2]
\end{split}
\end{equation}
were, $D^u$ is the discriminator, $\hat{x}^u =\alpha \psi(x^u)+(1-\alpha)\tilde{x}^u$, $\alpha\sim$U(0,1), and $\lambda$ is the gradient penalty coefficient. Note that unlike $D^s$, $D^u$ is not a conditional discriminator. Thus, the overall objective of the generator-discriminator pair for the unseen class instances can be defined as: 

\begin{equation}\label{eq:5}
    \underset{G^u}{\text{min}}\, \underset{D^u}{\text{max}}\,  \mathcal{L}^u_G
\end{equation}
The unlabeled class generator, $G^u$, trained in this fashion will produce synthetic unseen class latent representations that closely follow the true marginal distribution $P(\psi(x^u))$. However, it would not have learned the correct conditionals $P(\psi(x^u)|c(y^u))$. This is understandable as we do not have labeled unseen class data to train a conditional discriminator. On the other hand the seen class generator $G^s$ also models the $P(\psi(x^s)|c(y^s))$. This is because of the seen class conditional discriminator $D^s$. 

As the semantic label embeddings of the seen and unseen class share a common space and the latent representations of the seen and unseen class data are also from a common space, we hypothesize that generators of both the sets of classes must also be similar. Imposing this constraint allows us to transfer knowledge of the conditionals from the seen class generator to the unseen class generator.

%

Specifically, let $W^s_G$ be the weights associated with the seen class generator, $G^s$ and $W^u_G$ be the weights associated with the unseen class generator, $G^u$. We propose a weak transfer constraint that forces $W^u_G$ to be similar to that of $W^s_G$. We hypothesize that the unseen class generator learned using this constraint will encode the information on the conditionals.  Thus, the overall objective of the generator network for the unseen transfer is formulated as:    
\begin{equation}
    \underset{G}{\text{min}}\,  \underset{D}{\text{max}}\,  \mathcal{L}^u_{WGAN}+\omega\left\lVert W^s_G-W^u_G \right\rVert
\end{equation}
where, $\omega$  is a hyper-parameter controlling the importance of the similarity between the generators. When $\omega = 0$, the unseen class generator is completely independent of the seen class generator and there is no transfer of information between the two. This should result in synthetic unseen class instances that have very poor class conditional information in them. Large values of $\omega$ will force the unseen class generator to be identical to the seen class generator inducing high bias towards the seen classes; meaning the conditionals are biased towards the seen classes. This is also problematic as there is no overlap between the seen and unseen classes. Thus choosing an optimal hyper-parameter value that allows $G^u$ to learn from $G^s$ is important. This hyper-parameter is tuned through cross-validation on the set of seen classes.

\subsection{Classification and Evaluation Metric}
For the inductive setting, we generate synthetic unseen class representations using $G^s$. While for the transductive setting, we generate unseen class representations using $G^u$. As both $G^s$ and $G^u$ are conditional generators, we also have the unseen class labels associated with these synthetic instances. These are then used to train a softmax classifier to perform conventional ZSL. For the GZSL setting we combine the synthetically generated labeled  unseen class representations with the representations of seen class labeled training instances to learn a softmax classifier.


We average the correct predictions for each class and report the average per class top accuracy \cite{gbuxian2017zero,zerogbu} as below.
\begin{equation}
MCA_{u} = \frac{1}{|U|}\sum_{y^u \in U}{acc_{y^u}}
\end{equation}
\\
where, $acc_{y^u}$ denotes top-1 accuracy on test unseen data for each class in $U$.

In the generalized ZSL setting, we compute average per class top accuracy for both seen and unseen classes and report the harmonic mean of seen and unseen accuracy \cite{gbuxian2017zero,zerogbu} as defined below
\begin{equation}
    H = \frac{2*MCA_s*MCA_u}{MCA_s+MCA_u}
\end{equation}
where, $MCA_{u},MCA_{s}$ denote the mean class accuracy on test unseen and seen classes respectively.

\section{Experiments}

\subsection{Datasets}
We evaluate the proposed methods using the following three benchmark datasets of ZSL.
    \textbf {Animals with Attributes2 ({AWA2})} \cite{gbuxian2017zero,zerogbu} that comprises of 37,322 images belonging to 50 classes where each class label is described using a 85-dimensional vector. We use 40 classes for training and the remaining 10 classes for testing.
    \textbf{Caltech-UCSDBirds-200-2011 ({CUB})} \cite{cubdataset} that contains 11,788 images from 200 different types of birds where the class label is represented using a 312 dimensional vector. We use 150 classes for training and  the remaining 50 classes for testing.
    \textbf{SUN} \cite{sundataset} that consists of 14,340 images across 717 scenes with the class labels described using a 102 vector. Following \cite{gbuxian2017zero,zerogbu}, 580 classes out of 645 are used for training and the remaining 72 for testing. 

We employ the proposed splits of Xian \etal \cite{gbuxian2017zero,zerogbu} in all our experiments for fair comparison against prior approaches. For parameter tuning of the bias reducing models, we further divide the seen class set into train and validation splits preserving the original class distribution between train and test set. The best hyper-parameter values obtained through this cross-validation procedure is used to train the model using the entire training data.

\subsection{Experimental Settings}
\subsubsection{Learning the optimal latent space}
In all our experiments, we use Resnet-101\cite{resnet} as the deep convolutional architecture. {Prior literature has suggested using Resnet due to its on par or sometimes superior performance over other architectures such as VGG and Inception}. 
The transformation function $\psi$ is defined as two fully connected layers of 2048 and 1024 hidden units with ReLU activations acting on the final pooling layer of Resnet. The resulting 1024 dimensional space corresponds to the latent space, $\Psi$ of our network and is trained by the loss proposed in equation \ref{eq:3}. The classifier $f_c$ is defined as a single fully connected layer output layer with softmax activation over embedding space $\Psi$. The semantically aligning regressor $f_r$ is defined as a single output layer with the number of attributes in $c(y_s)$ over embedding space $\Psi$. This entire network is learned jointly. Learning rate is set to 0.001 and $\gamma$ is 0.01 across all the datasets. 

\subsubsection{Learning the bias reducing generators}
The seen and unseen class generators are modeled using a single fully connected hidden layer with 2048 leaky ReLU units. The discriminators contain a single hidden layer with 4096 leaky ReLU units that is connected to the output layer. Note that the SABR-I and SABR-T generators do not synthesize instances in the original data space, but the representations in the latent space $\Psi$. The gradient penalty coefficient of the Wasserstein generators, $\lambda$ is set to 10 and $\beta$, the coefficient that enforces the generators to learn the representations of the latent space is set to 0.1 across all datasets. We first train a conditional WGAN for the seen classes and then adapt the WGAN for the unseen classes. The rest of hyper-parameters are fine tuned by stratified cross-validation over the seen class training set.  
\section{Results and Discussion}
\subsection{Bias reduction in SABR-I}
\begin{figure}[t]
\includegraphics[width=0.47\textwidth]{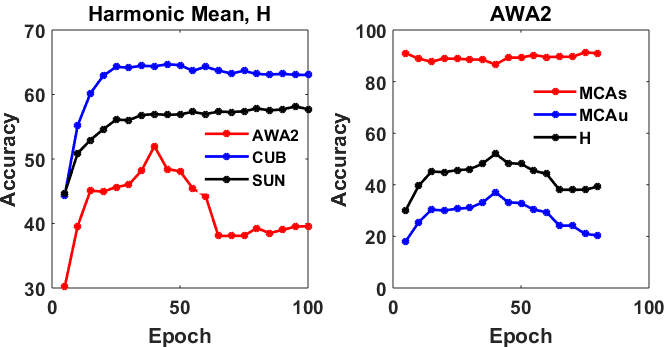}
\caption{[Best viewed in color] Effect of bias measured by plotting average harmonic mean, $H$ on the validation set (left). Detailed analysis on AWA2 by visualizing $MCA_s$, $MCA_u$ and $H$.}
\label{fig:biassabri}
\end{figure}

In the inductive setting, as we do not have the unseen class data, the generator $G^s$ tends to be overly biased towards the seen classes. We perform a three fold cross-validation experiment on the seen class training examples to validate this hypothesis. In each fold the seen class generator, $G^s$ was allowed to train for a long time till the generator and discriminator converged. Samples for the unseen class in the validation set were synthesized from the generator at different epochs. These instances were used to train a classifier model for the GZSL setting. The average harmonic mean, $H$ across the three folds as a function of the number of training epochs of the SABR-I generator is presented in Figure \ref{fig:biassabri}. It can be observed that as the number of training epochs increases, $H$ increases initially but starts decreasing beyond an epoch. For detailed analysis, we plot the $MCA_u$, $MCA_s$ and $H$ values for AWA2. We observe that as the training progresses, $MCA_s$ plateaus. However, the accuracy on the unseen classes in the cross-validation set $MCA_u$ increases initially, but starts decreasing beyond epoch 45. Due to this, the overall $H$ value decreases beyond this epoch. This behavior is very evident in the AWA2 dataset, and to a lesser extent in the CUB and SUN datasets due to significantly lesser number of training instances per class. This indicates that over training of the model based only on the seen class can bias the generator towards the seen classes and lose its capacity to generalize. This biasing is mitigated by stopping early at the best harmonic mean achieved \ie at the 40, 45, and 95 epoch for AWA2, CUB, and SUN datasets respectively.   
In the transductive setting, this bias is subdued due to the presence of unlabeled data and there is no need for early stopping.
\subsection{Bias reduction in SABR-T}
\begin{figure}[t]
\includegraphics[width=0.47\textwidth]{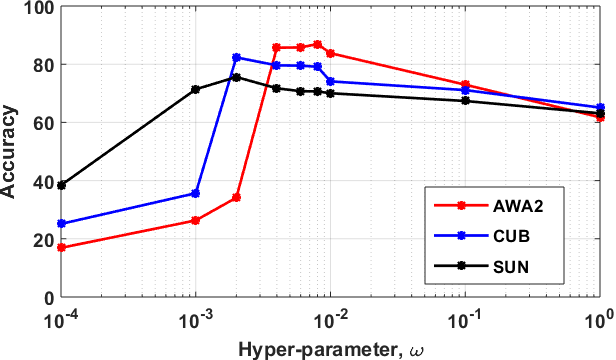}
\caption{[Best viewed in color] Effect of the hyper-parameter $\omega$ on the cross-validation set.}
\label{fig:biassabr-t}
\end{figure}

In the transductive setting, the hyper-parameter $\omega$ controls the bias in $G^u$ towards the seen classes. We tune this hyper-parameter through cross-validation where the seen classes are divided into seen and unseen classes, similar to the inductive setting. Figure \ref{fig:biassabr-t} illustrates the results obtained by the SABR-T while varying $\omega$ during the cross-validation process. We chose $\omega$ that performs best on this cross-validation set. For AWA2, CUB and SUN the model performs best at 0.008, 0.002 and 0.002 respectively. These values are fixed for the rest of the experiments.

\subsection{Performance in the conventional ZSL setting}
We compare our approach with existing state-of-the-art inductive and transductive ZSL methods. The results for the state of the art techniques reported in Table \ref{tab:1} are obtained from Xian \etal \cite{gbuxian2017zero,zerogbu} and use the proposed splits on AWA2, CUB, and SUN datasets with Resnet-101 as the underlying CNN architecture. 


\begin{table}[t]
\begin{center}
\begin{threeparttable}
    \begin{tabular}{|c| l| c| c| c|}
     \hline
     \textbf{Type}& \textbf{Method} & \textbf{AWA2} & \textbf{CUB} & \textbf{SUN}\\
     \hline
     & DAP\cite{lampert2014attribute}	& 46.1 &  40 & 39.9\\
    &IAP\cite{lampert2014attribute}	& 35.9	& 24 & 19.4\\
    &CONSE\cite{conse}	& 44.5	& 34.3	& 38.8\\
    &CMT\cite{cmtzerong}	& 37.9	& 34.6	& 39.9\\
    &SSE\cite{zhang2015zero}	&61	&43.9	&51.5\\
    &LATEM\cite{latem}	&55.8	&49.3	&55.3\\
    &ALE\cite{ale}	&62.5	&54.9	&58.1\\
    &DEVISE\cite{devise}	&59.7	&52	&56.5\\
    $I$&SJE\cite{sje}	&61.9	&53.9	&53.7\\
    &ESZSL\cite{embarrassinglyzsl}	&58.6	&53.9	&54.5\\
    &SYNC\cite{sync}	&46.6	&55.6	&56.3\\
    &SAE\cite{semanticae}	&54.1	&33.3	&4.3\\
    &GFZSL\cite{gfzsl}	&63.8	&49.4	&60\\
    &f-CLSWGAN\cite{DBLP:FeatureGenNetworks}	&-	&57.3	&60.8\\
    &PSR\cite{preservingsemantic}	&\textcolor{blue}{63.8}	&56	&\textcolor{blue}{61.4}\\
    &QFSL$^\#$\cite{qfzsl}	&63.5	& \textcolor{blue}{58.8}	&56.2\\
    \hline
    $I$&SABR-I  &\textbf{65.2}	& \textbf{63.9}	&\textbf{62.8}\\
    \hline
    \hline
    &ALE$^*$\cite{gbuxian2017zero}	&70.6	&{54.4}	&55.5\\
    $T$&GFZSL$^*$\cite{gfzsl}	&{78.3}	&50.6	&\textcolor{blue}{63.9}\\
    &DSRL$^*$\cite{dsrl}	&72.5	&48.9	&56.1\\
    &QFSL$^\#$\cite{qfzsl}	&\textcolor{blue}{79.7}	& \textcolor{blue}{72.1}	&58.3\\
    \hline
    
    $T$&SABR-T &\textbf{88.9}	& \textbf{74.0}	&\textbf{67.5}\\
    \hline
\end{tabular}
    \begin{tablenotes}
      \scriptsize	
      \item[1] $I$ - inductive ZSL setting, $T$- transductive ZSl setting.
      \item[2] Bold font denotes the best result  while blue refers to the second best result.
      \item[3] All results with $^*$ have been reported in Xian et al., \cite{gbuxian2017zero} and are optimistically approximated by their graphs.
      \item[$\#$] Note that results reported on QFSL are on GoogleNet architecture.
    \end{tablenotes}
    \caption{Performance in the conventional ZSL setting.}
    \label{tab:1}
    \end{threeparttable}
    \end{center}
\end{table}

The first observation from Table \ref{tab:1} is that no single state of the art method outperforms all the prior approaches on all the datasets for both the inductive and transductive settings. However, across all the inductive approaches, SABR-I performs better on AWA2 by 1.4\%, CUB by 5.07\% and by 1.3\% on SUN dataset. While in the transductive setting, SABR-T outperforms the current state-of-the-art by 9.2\%, 1.4\%, and 3.6\% on the AWA2, CUB and SUN datasets respectively. We attribute this high performance gain to the reduced bias in the model by the better alignment of the marginals and weak conditional transfer from the seen class generator. The next best performing model on AWA2 and CUB in the transductive setting, QFSL, assumes a uniform distribution over the unseen class label space, which we suspect is confusing the classifier.

The significant performance jump in the transductive setting over the inductive setting suggests the importance of having the marginal probability distribution of the unseen classes through the unlabeled instances. We further discuss the robustness of our approach with the amount of unlabeled data in section \ref{ablation}.
\begin{table*}[t]
\begin{center}
\begin{threeparttable}
    \begin{tabular}{|c| l| c| c| c|c| c| c|c| c| c|}
     \hline
    \textbf{Type} & \textbf{Method} & \multicolumn{3}{c|}{\textbf{AWA2}} &	\multicolumn{3}{c|}{\textbf{CUB}} &
    \multicolumn{3}{c|}{\textbf{SUN}}\\
    \hline
    & &  MCA$_u$&	MCA$_s$&	H&	MCA$_u$&	MCA$_s$&	H&	MCA$_u$&	MCA$_s$&	H\\
    \hline
    & DAP\cite{lampert2014attribute}&	0&	84.7&	0&	1.7&	67.9&	3.3&	4.2&	25.1&	7.2\\
    &IAP\cite{lampert2014attribute}&	0.9&	87.6&	1.8&	0.2&	72.8&	0.4&	1&	37.8&	1.8\\
    &CONSE\cite{conse}&	0.5&	90.6&	1&	1.6&	72.2&	3.1&	6.8&	39.9&	11.6\\
    &CMT\cite{conse}&	0.5&	90&	1&	7.2&	49.8&	12.6&	8.1&	21.8&	11.8\\
    &SSE\cite{zhang2015zero}&	8.1&	82.5&	14.8&	8.5&	46.9&	14.4&	2.1&	36.4&	4\\
    &LATEM\cite{latem}&	11.5&	77.3&	20&	15.2&	57.3&	24&	14.7&	28.8&	19.5\\
    &ALE\cite{ale}&	14&	81.8&	23.9&	23.7&	62&	34.4&	21.8&	33.1&	26.3\\
    &DEVISE\cite{devise}&	17.1&	74.7&	27.8&	23.8&	53&	32.8&	16.9&	27.4&	20.9\\
    $I$&SJE\cite{sje}&	8&	73.9&	14.4&	23.5&	59.2&	33.6&	14.7&	30.5&	19.8\\
    &ESZSL\cite{embarrassinglyzsl}&	5.9&	77.8&	11&	12.6&	63.8&	21&	11&	27.9&	15.8\\
    &SYNC\cite{sync}&	10.1&	90.5&	18&	11.5&	70.9&	19.8&	7.9&	43.3&	13.4\\
    &SAE\cite{semanticae}&	1.1&	82.2&	2.2&	7.8&	54&	13.6&	8.8&	18&	11.8\\
    &GFZSL\cite{gfzsl}&	2.5&	80.1&	4.8&	0&	45.7&	0&	0&	39.6&	0\\
    &f-CLSWGAN\cite{DBLP:FeatureGenNetworks}&	-&	-&	-&	43.7&	57.7&	\textcolor{blue}{49.7}&	42.6&	36.6&	\textcolor{blue}{39.4}\\
    &PSR\cite{preservingsemantic}&	20.7&	73.8&	\textcolor{blue}{32.3}&	24.6&	54.3&	33.9&	20.8&	37.2&	26.7\\
    \hline
    
    $I$ &SABR-I	&30.3	&93.9	&\textbf{46.9}	&55.03	&58.7	&\textbf{56.8}	&50.7	&35.1	&\textbf{41.5}\\
    \hline
    \hline
    &ALE$^*$\cite{gbuxian2017zero}&	 &		&21.7&	&		&30.4&	&	&	{21.1}\\
    $T$&GFZSL$^*$\cite{gfzsl}&	&		&		{40}	&		& &		{33.5}		&		& &		0\\
    &DSRL$^*$\cite{dsrl}&		&		&			32.3		&		& &			28.9		&		& &			20.5\\
    &QFSL\cite{qfzsl}&	66.2&	93.1&	\textcolor{blue}{77.4}&	71.5&	74.9&	\textbf{73.2}&	51.3&	31.2&	\textcolor{blue}{38.8}\\
    \hline
    $T$&SABR-T	&79.7	&91.0	&\textbf{85.0}	&67.2	&73.7	&\textcolor{blue}{70.3}	&58.8	&41.5	&\textbf{48.6}\\
    \hline
    \end{tabular}
    \begin{tablenotes}
      \scriptsize	
      \item[1] $I$ - inductive GZSL setting, $T$ - transductive GZSL setting
      \item[2] Bold font denotes the best result  while blue refers to the second best result.
      \item[3] All results with $^*$ only have H value reported by Xian et al., \cite{gbuxian2017zero} and are optimistically approximated by the graphs.
    \end{tablenotes}
    \caption{Performance in the Generalized ZSL Setting.}
    \label{tab:2}
\end{threeparttable}
\end{center}
\end{table*}
\subsection{Performance in the generalized ZSL setting}


Table \ref{tab:2} presents the results for the generalized ZSL setting for various state-of-the-art approaches. SABR-I improves over DAP\cite{lampert2014attribute}, IAP\cite{lampert2014attribute}, CONSE\cite{conse}, CMT\cite{cmtzerong}, SSE\cite{zhang2015zero}, LATEM\cite{latem}, ALE\cite{ale}, DEVISE\cite{devise}, SJE\cite{sje}, ESZSL\cite{embarrassinglyzsl}, SYNC\cite{sync}, SAE\cite{semanticae} and GFZSL\cite{gfzsl} by a large margin (over 10-20\%) across all the benchmark datasets. These methods utilized semantic space for the embedding and thus faced the hubness problem. We observe that SABR-I outperforms PSR\cite{preservingsemantic} by 14.65\%, 22.89\%, and 12.31\% on the AWA2, CUB and SUN datasets respectively.
This is because SABR-I not only preserves semantic relations in intermediate visual  space like PSR but also, subdues the  biasing effect by synthesizing unseen class instances and early stopping the training of the generator. SABR-I improves over f-CSLWGAN by 7.09\% on CUB dataset. We attribute this improvement to learning an optimal latent representation for the dataset that is fine tuned using the seen class data. However, on the SUN dataset the performance of SABR-I marginally better than f-CLSWGAN by 2.1\%. We speculate that the small number of training samples per class in the SUN dataset is insufficient to learn the optimal latent space and the seen class generator for SABR-I. 

In the transductive setting, SABR-T performs better than QFSL by 7.61\% and 9.38\% on the AWA2 and SUN datasets respectively. There are two potential reasons for this significant improvement. Firstly, QFSL fine-tunes the entire ResNet-101 architecture, while SABR-T learns a latent space that preserves both the semantic relations and discriminative information using the ResNet representations without fine tuning the ResNet model. This mitigates the hubness problem. Secondly, QFSL tries to reduce the bias of the seen classes in the latent space by encouraging the output for unseen class instances to be uniformly distributed across all the unseen classes. This results in loss of class conditional information in the latent space. In contrast, SABR-T aligns the marginal distribution of the generated samples with that of the true unseen class instances, and learns the class conditionals from the seen class data. On the CUB dataset, while SABR-T yields superior performance over previous transductive approaches by 36.31\%, except QFSL, where there is a marginal drop of 2.9\%. 

\begin{figure}[t]
\includegraphics[width=0.49\textwidth]{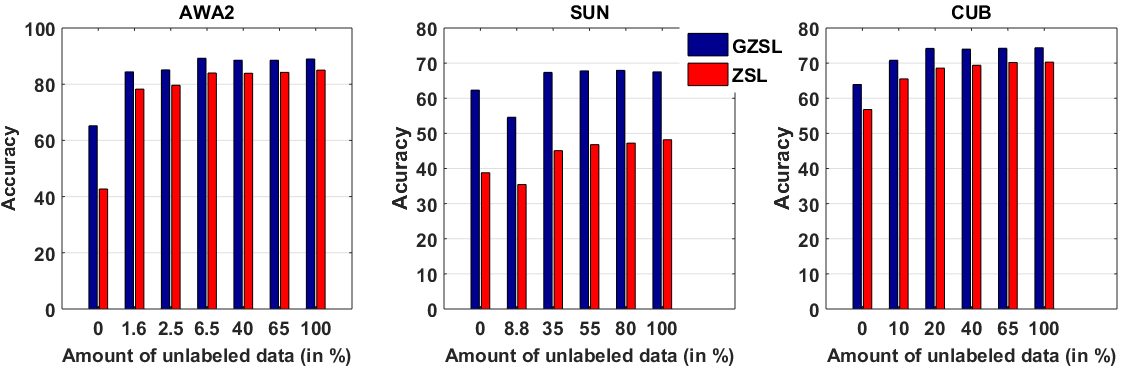}
\caption{Change in the transductive ZSL and GZSL accuracy as a function of the number of unseen class instances.}
\label{fig:4}
\end{figure}
\subsection{Amount of unlabeled data for transductive learning}\label{ablation}
Prior transductive ZSL approaches and SABR-T utilize all the available unlabeled unseen class instances for modeling the marginal probability distributions in both ZSL and GZSL settings.  We hypothesize that the marginals of the unseen class instances can be learned from a smaller subset of unlabeled data.  We conduct experiments using SABR-T on the AWA2 and SUN datasets as these contain maximum and minimum number of instances per class. SABR-T is trained with a randomly selected subset of unlabeled unseen class instances and the average performance over 5 trials for both ZSL and GZSL settings are reported in Figure \ref{fig:4}. As expected the performance on both the datasets increases with increase in the number of unlabeled instances used during training. Interestingly, on the AWA2 dataset, SABR-T achieves the best performance with using only 6.5\% of the total unlabeled instances. We observe a similar trend on the CUB and SUN dataset with the performance peaking by using only 10\% and 35\% of the unlabeled data respectively.

\section{Summary and Future Work}
 In this work, we propose a novel approach, Semantically Aligned Bias Reducing (SABR) Zero Shot Learning, which focuses on solving both the hubness and bias problems that are commonly faced by zero shot learning (ZSL) techniques. SABR overcomes the hubness problem by learning a latent space that preserves the semantic relationship between the labels while encoding the discriminating information about the classes. Further, we also propose ways to reduce bias of the seen classes through a simple cross-validation process in the inductive setting and a novel weak transfer constraint in the transductive setting.  Extensive experiments are conducted on three benchmark datasets (AWA2, CUB, and SUN) to investigate the efficacy of the proposed model. The results suggest that SABR significantly outperforms existing state-of-the-art zero shot learning algorithms by $\sim$1.5-9\% in the conventional ZSL setting and by $\sim$2-14\% in the generalized ZSL for both the inductive and transductive settings on most of the datasets. We also demonstrate that SABR reaches peak performance in the transductive setting by using only a fraction of the unlabeled unseen class instances for training.

In future, we would like to explore other semantic spaces like word vector embeddings or combination of different semantic label embeddings to accurately model the relationship between the seen and unseen class labels. It would also be beneficial to explore ways to extract semantic label embeddings from auxiliary sources as this would assist in extending ZSL to many real-world scenarios.

\section{Acknowledgement}
We  would  like  to  thank  the  anonymous  reviewers  for their suggestions. We are grateful to V. Karthik for the discussions during the early phase of this work and NVIDIA for supporting this work through the academic hardware grant program.
{\small
\bibliographystyle{ieee_fullname}
\bibliography{egbib}
}

\end{document}